\documentclass[letterpaper]{article} 
\usepackage{aaai2026}
\usepackage{times}  
\usepackage{helvet}  
\usepackage{courier}  
\usepackage[hyphens]{url}  
\usepackage{graphicx} 
\usepackage{natbib}  
\usepackage{caption} 
\usepackage{algorithm}
\usepackage[most]{tcolorbox} 

\usepackage{algorithmic}

\usepackage{newfloat}
\usepackage{listings}
\DeclareCaptionStyle{ruled}{labelfont=normalfont,labelsep=colon,strut=off} 
\floatstyle{ruled}
\newfloat{listing}{tb}{lst}{}
\floatname{listing}{Listing}
\title{AdEval: Alignment-based Dynamic Evaluation to Mitigate Data Contamination in Large Language Models}
\author{
    Fan Yang
}
\affiliations{
    \textsuperscript{\rm }Jinan University, Guangzhou, China, winstonyf@qq.com\\

}

\usepackage{bibentry}

\begin{document}

\maketitle

\begin{abstract}
As Large Language Models (LLMs) are pre-trained on ultra-large-scale corpora, the problem of data contamination is becoming increasingly serious, and there is a risk that static evaluation benchmarks overestimate the performance of LLMs. To address this, this paper proposes a dynamic data evaluation method called AdEval (Alignment-based Dynamic Evaluation). AdEval first extracts knowledge points and main ideas from static datasets to achieve dynamic alignment with the core content of static benchmarks, and by avoiding direct reliance on static datasets, it inherently reduces the risk of data contamination from the source. It then obtains background information through online searches to generate detailed descriptions of the knowledge points. Finally, it designs questions based on Bloom's cognitive hierarchy across six dimensions-remembering, understanding, applying, analyzing, evaluating, and creating to enable multi-level cognitive assessment. Additionally, AdEval controls the complexity of dynamically generated datasets through iterative question reconstruction. Experimental results on multiple datasets show that AdEval effectively alleviates the impact of data contamination on evaluation results, solves the problems of insufficient complexity control and single-dimensional evaluation, and improves the fairness, reliability and diversity of LLMs evaluation.

\end{abstract}

\begin{links}
    \link{Code}{https://anonymous.4open.science/r/AdEavl-F794}
\end{links}

\begin{figure*}[htbp]
  \centering
  \includegraphics[width=0.9\linewidth]{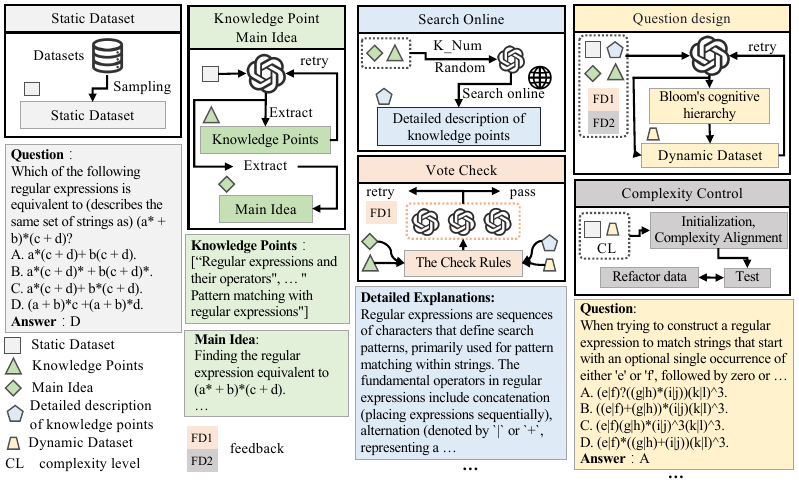}
  \caption {An Overview of the AdEval Process: Generating Dynamic Datasets from Static Data.}
  \label{fig:AdEval}
\end{figure*}

\section{Introduction}

Large Language Models (LLMs) have significantly advanced the field of Natural Language Processing (NLP) with their remarkable performance across a wide range of tasks~\cite{achiam2023gpt,yang2025qwen3}. 
LLMs rely on pretraining over massive-scale internet corpora, while many widely-used benchmarks are also derived from online resources, which inevitably leads to the issue of data contamination, i.e., the phenomenon that examples from the evaluation set are also found in the training data \cite{achiam2023gpt,deng2024unveiling,zhu2024dyval,li2024open,yu2024kieval}. As LLMs grow in scale, overlaps between training data and evaluation benchmarks have become increasingly common, which will lead to misunderstandings of LLMs' performance and affect the fairness and validity of LLMs comparisons.

Several studies have investigated the impact of data contamination. ~\citet{magar2022data} and \citet{jiang2024investigating} explored how data contamination during pretraining affects language model performance, analyzing factors such as data volume and model size. \citet{palavalli2024taxonomy} and  \citet{li2024task} examined the influence of contamination on downstream tasks or zero- and few-shot evaluations. To address data contamination, two primary solutions have been proposed: data contamination detection and dynamic data evaluation \citep{yu2024kieval}.

Data contamination detection aims to identify overlaps between model outputs and training data \citep{balloccu2024leak}. While various methods have been proposed—such as sample swapping, perplexity-based approaches, and analysis of internal model states \citep{dong2024generalization,dekoninck2024constat,golchin2024time,yao2024data,maini2024llm,oren2023proving,ye2024data,ni2025training,shi2024detecting}—many rely on intrinsic information from LLMs or complex analytical techniques. However, some closed-source models (such as the GPT series) incorporate specific filtering mechanisms during generation, which prevent them from directly reproducing training data, thereby limiting the effectiveness of detection methods. Moreover, existing detection methods typically focus on extreme memorization (i.e., verbatim reproduction), which makes them less effective at identifying subtler forms of contamination. Complicating matters further, the training data of closed models is often considered proprietary and is not publicly accessible, making it difficult to verify contamination even when internal signals are available. This lack of transparency significantly reduces the practicality of detection approaches that depend on access to training data or model internals.


Compared with data contamination detection, dynamic data evaluation can alleviate the data contamination problem to a large extent. However, existing dynamic evaluation methods still have some limitations. For example, DYVAL \cite{zhu2024dyval} and SciEval \cite{sun2024scieval} dynamically generate data for the mathematics field and the scientific research field respectively, but both are limited to specific fields and cannot support the evaluation of the general capabilities of LLMs. KIEval \cite{yu2024kieval} introduced an evaluation mechanism of multi-round dialogues, but the accuracy of answers cannot be guaranteed, which affects the reliability of the evaluation results. LatestEval \cite{li2024latesteval} dynamically generates the reading comprehension dataset through web search, but the generated data has the problems of high randomness and being too simplified. \citet{ying2024automating} update the dataset through strategies of imitating and expanding existing datasets, however, the generated dataset has high similarity to the original dataset, and there is still a risk of data contamination. The existing dynamic evaluation mechanisms have not yet achieved the unified consideration of domain universality, evaluation reliability, complexity controllability and anti-contamination.


To this end, this paper proposes a dynamic data evaluation method called AdEval (Alignment-based Dynamic Evaluation), which dynamically constructs evaluation datasets by aligning with static benchmarks. By avoiding direct reliance on static datasets, AdEval effectively reduces the risk of data contamination. Moreover, it incorporates multi-level cognitive assessment based on Bloom’s taxonomy hierarchy and manages complexity through iterative data reconstruction. The contributions of AdEval are as follows:

\begin{enumerate}

    \item AdEval performs dynamic alignment by extracting and expanding core content from static datasets, minimizing the risk of data contamination from the source.
    \item AdEval controls the complexity of dynamically generated datasets relative to static datasets automatically through multiple rounds of question reconstruction.
    
    \item AdEval evaluates LLMs from multiple dimensions through Bloom's cognitive hierarchy, not only assessing basic abilities, but also emphasizing the examination of LLMs' high-level cognitive abilities.
\end{enumerate}

\section{Related Work}

\subsection{Benchmarks for Evaluating Large Language Models}
There are many static evaluation benchmarks for evaluating LLMs. For example, MMLU \cite{hendrycks2020measuring} assesses the knowledge and reasoning abilities of models in multiple disciplines. HELM \cite{liang2022holistic} expands evaluation dimensions such as fairness and efficiency. BIG-Bench \cite{srivastava2022beyond} tests the limit of model capabilities through 204 challenging tasks. AGIEval \cite{zhong2024agieval} and C-Eval \cite{huang2023c} focus on the evaluation of Chinese knowledge. ARC-Challenge \cite{clark2018think} evaluates the scientific knowledge and reasoning abilities of models, and BELLE \cite{ji2023exploring} evaluates the understanding and generation abilities of models in natural language processing.

\subsection{Data Contamination}


Dynamic data evaluation can alleviate the data contamination problem to a large extent. LatestEval \cite{li2024latesteval} and \citet{ying2024automating} dynamically generate datasets based on the latest text; DYVAL \cite{zhu2024dyval} and DyVal 2 \cite{zhu2024dyval} dynamically generate datasets with controllable complexity; SciEval \cite{sun2024scieval} combines static and dynamic data to generate a dataset for multi-level evaluation of scientific capabilities. \citet{liu2024evaluating} generate a dataset for comprehensively examining Chinese knowledge capabilities. Clean-Eval \cite{zhu2024clean} optimizes data by using paraphrasing and semantic detection.  Mehrbakhsh et al. \cite{mehrbakhsh2024confounders} generate datasets by creating variant and restated questions. \citet{yang2023rethinking} filters similar samples to reduce the risk of overfitting. KIEval \cite{yu2024kieval} evaluates the capabilities of LLMs through multi-round dialogues. These studies provide diverse strategies for dynamic evaluation.

\section{Methods}

The process of dynamically generating datasets proposed in this paper, as shown in Figure~\ref{fig:AdEval}, consists of the following steps: extraction of knowledge points and main ideas, online search, question design, complexity control, and vote check.

\subsection{Knowledge Point and Main Idea Extraction}

First, extract the corresponding knowledge points from the sampled dataset. During the extraction process, the knowledge points are generated autonomously by LLMs and must meet the following three requirements: (1) the generated knowledge points must be relevant to the question; (2) focus on summarizing broad concepts or knowledge categories related to the question; (3) the generated format must conform to the predefined requirements.

Then, extract the main idea from the question. The LLMs generate the main idea by summarizing the question, which must meet the following two requirements: (1) the main idea should cover the background of the question and the core of the answer; (2) the main idea should be concise and clear. The main idea plays an important role in the subsequent online search and question design steps, ensuring that the newly generated questions align with the core concepts of the static data questions.

In summary, the knowledge points are used to control the different directions of the questions, while the main idea is used to maintain the core concepts of the questions. An example is as follows:

\begin{tcolorbox}[
    colback=gray!00,
    colframe=black,
    arc=1.5mm, 
    auto outer arc, 
    left=0.9mm, 
    right=0.9mm, 
    boxrule=0.9pt, 
    top=0mm,  
    bottom=0mm,  
    title={Extract Knowledge Point and Main Idea},
    fonttitle=\centering\fontsize{9pt}{9pt}\selectfont, 
    fontupper=\fontsize{8pt}{9pt}\selectfont 
]
Question: Which of the following regular expressions is equivalent to (describes the same set of strings as) (a* + b)*(c + d)?

A. a*(c + d)+ b(c + d). B. a*(c + d)* + b(c + d)*

C. a*(c + d)+ b*(c + d). D. (a + b)*c +(a + b)*d.

Answer: D

Knowledge Point: ["Regular expressions and their operators", "Concatenation in regular expressions", "Union operator (+) in regular expressions", "Kleene star (*) in regular expressions"...]

Main Idea: Finding the regular expression equivalent to (a* + b)*(c + d).
\end{tcolorbox}

\subsection{Online Search for detailed descriptions of knowledge points}

Using the Online Search function of LLMs, AdEval takes the question stems, knowledge points and their corresponding main ideas of each question in the dataset as input, and generates detailed descriptions focusing on the knowledge points based on relevant background information such as the latest materials, research results, definitions, and cases retrieved online.

\begin{tcolorbox}[
    colback=gray!00,
    colframe=black,
    arc=1.5mm, 
    auto outer arc, 
    left=0.9mm, 
    right=0.9mm, 
    boxrule=0.9pt, 
    top=0mm,  
    bottom=0mm,  
    title={Online Search Example},
    fonttitle=\centering\fontsize{9pt}{9pt}\selectfont, 
    fontupper=\fontsize{8pt}{9pt}\selectfont 
]
Regular expressions are sequences of characters that define search patterns, primarily used for pattern matching within strings. The fundamental operators in regular expressions include concatenation (placing expressions sequentially), alternation (denoted by `|` or `+`, representing a choice between expressions), and the Kleene star (denoted by `*`, indicating zero or more repetitions of the preceding element)… 
\end{tcolorbox}

Each question is associated with one main idea and multiple knowledge points. In actual tests, this paper finds that generating knowledge point explanations takes much time. To improve the flexibility of generation, this paper designs a knowledge point selection mechanism, which controls the number of selected knowledge points through parameters, namely: randomly select a specified number of knowledge points from the knowledge point array for processing. For example, when the parameter is set to 2, if the number of knowledge points exceeds 2, 2 are randomly selected; if the number of knowledge points is less than 2, all knowledge points are selected.

\subsection{Question Design}

In this section, by embedding static data questions, knowledge points, main ideas, and detailed descriptions of knowledge points into the prompts, two types of new questions are generated: regular new questions and Bloom's cognitive hierarchy-based new questions \cite{krathwohl2002revision,forehand2010bloom}. The generated questions must meet the following requirements: high relevance, consistent difficulty with static questions, reasonable and distinguishable options, avoiding external dependencies, unique answers, assessing deep understanding, innovative question stems, different emphasis, and focusing on knowledge points. Below is an example of a new question based on Bloom's cognitive hierarchy:

\begin{tcolorbox}[
    colback=gray!00,
    colframe=black,
    arc=1.5mm, 
    auto outer arc, 
    left=0.9mm, 
    right=0.9mm, 
    boxrule=0.9pt, 
    top=0mm,  
    bottom=0mm,  
    title={Question Design Example},
    fonttitle=\centering\fontsize{9pt}{9pt}\selectfont, 
    fontupper=\fontsize{8pt}{9pt}\selectfont 
]
Based on Bloom's cognitive hierarchy

Question: Which symbol in regular expressions denotes zero or more repetitions of the preceding element? 

 A: ?     B: *     C: +     D: .

Answer: B, Layer: Remembering

...

 Question: If you were to create a new regular expression based on the pattern `(a* + b)*(c + d)` that matches a string starting with zero or more 'a's, followed by one or more 'b's, and ending with either 'c' or 'd', what would that regular expression be?

A: (a*b+)(c|d) B: (a+b*)(c|d)

C: (a* + b+)(c|d) D: (a+b*)(c+d)

Answer: C, Layer: Creating
\end{tcolorbox}

\subsection{Question Complexity Control}

\begin{figure}[htbp]
    \centering
  \includegraphics{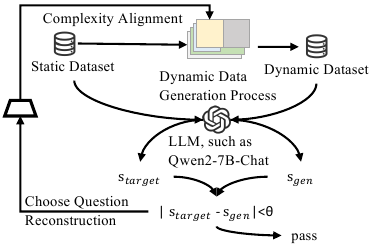}
  \caption{Question Complexity Adjustment Process - Aligning with Static Dataset Example}
  \label{fig:QuestionDifficultyAdjustment}
\end{figure}

If no constraints are imposed on the process of generating questions by LLMs, the results tend to generate relatively simple dataset questions. This paper designs a question complexity adjustment process, as shown in Figure~\ref{fig:QuestionDifficultyAdjustment}, to ensure that the complexity of the generated dataset aligns with that of the static dataset.

This paper indirectly measures the complexity of the dataset by the precision when tested on LLMs, such as Qwen-7B-Chat \cite{yang2025qwen3}. The newly generated dataset will be input into the LLM for testing, the model's precision will be calculated, and the offset of the question complexity will be obtained based on this: $\Delta = s_{\text{target}} - s_{\text{gen}}$, Where \( s_{\text{target}} \) represents the desired precision, and \( s_{\text{gen}} \) represents the precision of the newly generated dataset on the LLM. It aims for the deviation to satisfy \( |\Delta| < 5\% \).

Adjustment Strategy: If \( \Delta > 5\% \): The generated dataset’s complexity is too high. We then select the questions that the LLM answered incorrectly from the generated dataset and reconstruct them to simplify the questions;  If \( \Delta < -5\% \): The generated dataset’s complexity is too low. We then select the questions that the model answered correctly and reconstruct them to increase the complexity. Repeat the above steps until the newly generated dataset satisfies \( |\Delta| < 5\% \). 

\subsection{Vote Check}

In terms of question quality check, manual checking is not only time-consuming but also requires a lot of energy. For this reason, many researchers have proposed using LLMs for evaluation and judgment. For example, \citet{wei2023jailbreak} uses a language model (fine-tuned on Llama-13b) as an evaluation tool. \citet{liu2023autodan} uses the GPT Recheck tool. \citet{qi2024fine} uses a GPT-4 Judge to provide a more contextualized assessment. The GPT-4 Judge assesses harmfulness on a 1 to 5 Likert scale (1=least harmful, 5=most harmful). However, in our past work, it was found that using a single LLM as an evaluation tool has certain instability. In order to ensure the quality of the questions generated by AdEval, this paper adopted a multi-model voting check mechanism, using multiple LLMs for quality voting evaluation. The voting check rules are as follows: if more than half of the answers from multiple LLMs judge as "pass", the question is considered to have passed the quality check; if this condition is not met, the question is deemed to have quality issues and will be regenerated. Specific quality issues include: generating questions that are too similar to the original questions, unclear logic, unreasonable options, wrong answers, failure to cover core knowledge points, etc.

\subsection{Algorithm Description  }

The algorithm of AdEval proposed in this paper is shown in Algorithm \ref{alg:AdEval}

\begin{algorithm}[tb]
\fontsize{9pt}{11pt}\selectfont
\caption{AdEval}
\label{alg:AdEval}
\textbf{Input}: Static evaluation dataset $D_{\mathrm{static}}$, number of dataset samples $S$, number of knowledge points $K_{\mathrm{num}}$, complexity alignment threshold $\varepsilon$, Bloom's cognitive hierarchy $Bloom$, Large Language Model $LLM$ \\
\textbf{Output}: Dynamic evaluation dataset $D_{\mathrm{dynamic}}$

\begin{algorithmic}[1]
\STATE Initialize knowledge points $KN$, main idea $m_{idea}$, and array of knowledge point explanations $KNexplain$
\STATE ${D'}_{\mathrm{static}} \leftarrow \left\{ d_{\left\lfloor\frac{k(N-1)}{S}\right\rfloor} \mid k=0,1,\dots,S-1 \right\},\ \ N = \text{len}(D_{\mathrm{static}})$
\FOR{each $d_i$ in ${D'}_{\mathrm{static}}$}
    \STATE ${kn}_i \leftarrow LLM(d_i)$
    \STATE $n \leftarrow \text{len}({kn}_i)$
    \STATE $KN \leftarrow 
    \begin{cases}
        \text{rand\_select}({kn}_i, K_{\text{num}}), & \text{if } n \geq K_{\text{num}} \\
        {kn}_i, & \text{otherwise}
    \end{cases}$
    \STATE $m_i \leftarrow LLM(d_i)$
    \FOR{each ${kn}_{i_j}$ in ${kn}_i$}
        \STATE $e_{i_j} \leftarrow LLM_{\text{net}}(d_i, {kn}_{i_j}, m_i)$
        \STATE $Q_{i_j} \leftarrow LLM(d_i, {kn}_{i_j}, m_i, e_{i_j}, Bloom)$
        \STATE Add $Q_{i_j}$ to $D_{\mathrm{dynamic}}$
    \ENDFOR
\ENDFOR
\STATE $\Delta \leftarrow LLM_{\text{test}}(D_{\mathrm{dynamic}}) - LLM_{\text{test}}({D'}_{\mathrm{static}})$
\WHILE{$|\Delta| < \varepsilon$}
    \STATE $Q_{\text{error}} \leftarrow Q\_select\_error(LLM_{\text{test}}(D_{\mathrm{dynamic}}))$
    \STATE $Q_{\text{correct}} \leftarrow Q\_select\_correct(LLM_{\text{test}}(D_{\mathrm{dynamic}}))$
    \STATE $D_{\mathrm{dynamic}} \leftarrow 
    \begin{cases}
        \text{restructure}(Q_{\text{error}}, \text{simple}), & \text{if } \Delta > 0 \\
        \text{restructure}(Q_{\text{correct}}, \text{complex}), & \text{if } \Delta > 0
    \end{cases}$
    \STATE $\Delta \leftarrow LLM_{\text{test}}(D_{\mathrm{dynamic}}) - LLM_{\text{test}}({D'}_{\mathrm{static}})$
\ENDWHILE
\FOR{each $d_i$ in $D_{\mathrm{dynamic}}$}
    \IF{$Multi\_LLM(d_i) == \text{pass}$}
        \STATE \textbf{continue}
    \ELSE
        \STATE $d_i \leftarrow \text{restructure}(d_i)$
    \ENDIF
\ENDFOR
\STATE \textbf{return} $D_{\mathrm{dynamic}}$
\end{algorithmic}
\end{algorithm}

In the AdEval algorithm, there are three main modules: the dynamic dataset generation module, the complexity control module, and the quality check module. In Steps 2 and 6 of the algorithm, the size of \( D_{\text{dynamic}} \) is controlled by the sampling number \( S \) and the number of knowledge points \( K_{\text{num}} \). The notations \( LLM \), \( \text{LLM}_{\text{net}} \), \( \text{LLM}_{\text{test}} \), and \( \text{Multi\_LLM} \) in the algorithm represent the large language model, the large language model with internet access, using the large language model for precision testing, and using multiple large language models for checking, respectively. The function \( \text{restructure}(Q, \text{simple/complex}) \) in the algorithm denotes using the large model to restructure the question \( Q \), where \( \text{simple} \) indicates restructuring toward simplicity and \( \text{complex} \) indicates restructuring toward complexity. If \( \text{simple/complex} \) is not specified, it means normal restructuring without any bias.

\section{Experiment}

In this section, this paper conducted experimental verification on AdEval: (1) to evaluate the impact of data contamination on AdEval; (2) to evaluate AdEval's ability to align with the complexity of static datasets; (3) to evaluate AdEval's ability to generate multi-dimensional questions; (4) to evaluate the quality of data generated by AdEval.

\subsection{Experiment Setup}

\textbf{Dataset}: The datasets used in the experiment include MMLU and ARC-Challenge, 300 samples were sampled from each of these datasets. Additionally, there is the Question-Answering(QA) dataset BELLE. To reduce the interference of randomness on the results, each group of data was independently evaluated five times, and the inference temperature of LLMs for the data was set to 0.

\textbf{Few-Shot}: AdEval adopts the Few-Shot learning strategy. For knowledge point extraction, main idea extraction, explanations generation, and question generation, 2-shot is used. Due to the excessively long context length, 1-shot is used for question generation based on Bloom's cognitive hierarchy.

\textbf{Model}: The LLMs used in the experiment are as follows: GPT4o \cite{achiam2023gpt}, GPT-3.5-turbo \cite{brown2020language}, DeepSeek-V3 \cite{guo2025deepseek,liu2024deepseek} (671B), glm-4-flash \cite{glm2024chatglm}, qwen-plus \cite{hui2024qwen2}, Claude-3-5-haiku, Doubao-pro-32k, Qwen2-7B-Chat \cite{hui2024qwen2} (7B), Llama3-8B-Instruct \cite{dubey2024llama} (8B), glm-4-9b-chat (9B). 

\textbf{Fine-Tuning}: Fine-tuning used the LoRA method with the following configuration: learning rate set to \( 5 \times 10^{-5} \), number of training epochs set to 2, batch size set to 2, LoRA rank set to 8, scaling factor set to 16, dropout rate set to 0.1, and LoRA+ learning rate ratio set to 16.

\subsection{Evaluation Metrics}

\subsubsection{Similarity Metric for QA Tasks}  
In the QA task, we compute the similarity between the model's outputs before and after fine-tuning using the Longest Common Subsequence (LCS) method to measure string similarity. The formula is as follows:

\begin{equation}
  \label{eq:Similarity}
  \text{Similarity} = \frac{\text{LCS}(X, Y)}{\max(|X|, |Y|)} 
\end{equation}

where \(X\) and \(Y\) represent the two input texts. \(\left|X\right|\) and \(\left|Y\right|\) represent the number of characters in \(X\) and \(Y\) respectively (excluding spaces and punctuation). \(\max(\ )\) means taking the maximum value among the numbers, \(LCS(X,Y)\) represents the length of the longest common subsequence of texts \(X\) and \(Y\), and finally normalizing to make the similarity between 0 and 1.

\subsubsection{Perplexity}

Perplexity measures the "uncertainty" of the content generated by the LLM. The perplexity formula for the dataset is as follows:

\begin{equation} 
  \label{eq:equation6}
  \text{PPL}_{dataset} = \frac{1}{M} \sum_{j=1}^{M} 2^{-\frac{1}{N}\sum_{i=1}^{N}\log_2(p_i)}
\end{equation}

\(M\) represents the total number of samples. \(\log_2(p_i)\) is the log-probability of the \(i\)-th token generated in the \(j\)-th sample. \(N\) is the number of tokens in each sample. \(\frac{1}{N}\sum_{i = 1}^{N}\log_2(p_i)\) is the average log-probability of the \(j\)-th sample in the dataset.

\begin{figure}[htbp]
  \centering
  \includegraphics[width=\columnwidth]{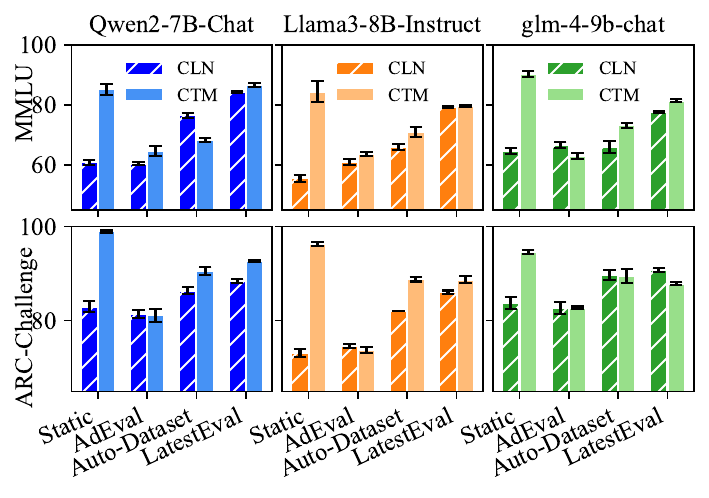}
  \caption {Based on the MMLU and ARC-Challenge datasets, new datasets are dynamically generated on AdEaval, Auto-dataset, and Latesteval respectively. Then, precision comparisons of these new datasets are conducted on Qwen-7B-Chat, Llama3-8B-Instruct, and glm-4-9b-chat before and after contamination.}
  \label{fig:Data_Contamination_Fine-Tuning_combined}
\end{figure}

\subsection{Data Contamination Test}

In this experiment, the static dataset is used to fine-tune small-parameter LLMs with data contamination.   When examining the impact of data contamination on the fine-tuning effect of small-parameter LLMs, two models are used for comparison: one is the clean model (CLN, Clean), and the other is the model contaminated by static data (CTM, Contamination).   The experimental results are compared and analyzed through different datasets to quantify the impact of static data contamination and dynamic data generation methods on model performance.

As shown in Figure \ref{fig:Data_Contamination_Fine-Tuning_combined}, AdEval represents the dynamic data generated by the AdEval method, Auto-Dataset represents the dynamic data generated by Auto-Dataset, and LatestEval represents the dynamic data generated by LatestEval.   Figure \ref{fig:Data_Contamination_Fine-Tuning_combined} has two rows: the first row is the experiment based on the MMLU dataset, and the second row is the experiment based on ARC-Challenge.   The three columns represent the experiments based on Qwen-7B-Chat, Llama3-8B-Instruct, and glm-4-9b-chat respectively.   Each subgraph shows the accuracy of LLMs before and after contamination between the static dataset and the datasets generated by AdEval, Auto-Dataset, and LatestEval.   It can be seen from the first group of static data histograms (Static) in each subgraph of Figure  \ref{fig:Data_Contamination_Fine-Tuning_combined} that data contamination will falsely increase the accuracy of large models, improving the precision by more than 20\%.   However, data contamination has little impact on dynamically generated data.   Compared with LatestEval and Auto-Dataset, AdEval has the smallest impact on contaminated models, and only AdEval can maintain basically the same complexity as static data.

\subsection{Perplexity Test}
\label{sec:Perplexity Test}
In this experiment, perplexity tests were conducted on the static data and three types of dynamically generated data  on Qwen-7B-Chat before and after data contamination.

\begin{table}[htbp]
  \centering
  \small 
  \setlength{\tabcolsep}{3pt} 
  \begin{tabular}{c|ccc|ccc}
    \hline
    & \multicolumn{3}{c}{\textbf{MMLU \%}} & \multicolumn{3}{|c}{\textbf{ARC-Challenge \%}} \\ 
    & CLN & CTM & $\Delta$ & CLN & CTM & $\Delta$ \\ \hline
    Static & 1.31 & 1.14 & \textit{\textcolor{magenta}{0.17$\downarrow$}} & 1.24 & 1.01 & \underline{\textcolor{orange}{0.23$\downarrow$}} \\ 
    AdEaval & 1.32 & 1.23 & \textbf{\textcolor{blue}{0.09$\downarrow$}} & 1.30 & 1.29 & \textbf{\textcolor{blue}{0.01$\downarrow$}} \\ 
    Auto-dataset & 1.22 & 1.16 & \textbf{\textcolor{blue}{0.06$\downarrow$}} & 1.21 & 1.03 & \textit{\textcolor{magenta}{0.18$\downarrow$}} \\ 
    Latesteval & 1.20 & 1.07 & \textit{\textcolor{magenta}{0.13$\downarrow$}} & 1.20 & 1.06 & \textit{\textcolor{magenta}{0.14$\downarrow$}} \\ 
    \hline
  \end{tabular}
  
  \caption{Based on the MMLU and ARC-Challenge datasets, new datasets are dynamically generated respectively on AdEaval, Auto-dataset, and Latesteval. Then, perplexity test comparisons are carried out on Qwen-7B-Chat before and after contamination.}
  \label{tab:Perplexity}
\end{table}

In Table~\ref{tab:Perplexity}, The $\Delta$ value represents the difference in LLM perplexity before and after contamination. \underline{\textcolor{orange}{$\Delta > 0.2$ is marked with an underline}}, \textit{\textcolor{magenta}{$\Delta \in [0.1,0.2]$ is marked in italics}}, and \textbf{\textcolor{blue}{$\Delta \in [0,0.1]$ is marked in bold}}. 

From Table~\ref{tab:Perplexity}, it can be seen that for static data, the perplexity of LLMs before and after contamination decreases significantly, with a $\Delta$ value around 0.2. In dynamic dataset, the $\Delta$ values of AdEval are all below 0.1, with a small decrease and stable performance. In contrast, the $\Delta$ values of LatestEval range from 0.1 to 0.2, and the $\Delta$ value of Auto-Dataset is between 0.05 and 0.2, showing significant fluctuations. As can be seen from the table, both static data and dynamic data exhibit a decrease in perplexity when tested on LLMs. This is because the dynamically generated dataset belongs to the same question type as the static data. Although the questions differ, fine-tuning makes the model more familiar with the question type, resulting in an overall decrease in perplexity.

\subsection{Question-Answering Similarity Test}
\label{sec:Question-Answering Similarity Test}


To evaluate the impact of data contamination on the LLMs in the QA task, this experiment quantified the overlap degree of the output texts of the LLMs before and after contamination by calculating the Longest Common Subsequence (LCS) similarity between the two sets of answers. Specifically, the BELLE dataset is used to test the similarity of the answers of Qwen-7B-Chat before and after contamination. The results are shown in Table~\ref{tab:Similarity}.


\begin{table}[htbp]
  \centering
  \small 
  \begin{tabular}{c|ccc}
    \hline
    & \multicolumn{3}{c}{\textbf{LCS \%}} \\ 
    BELLE & CLN & CTM & $\Delta$ \\ \hline
    Static & 17.57 & 42.17 & 24.6$\uparrow$ \\ 
    AdEval & 18.63 & 18.65 & 0.02$\uparrow$  \\
    \hline
  \end{tabular}
  		
  \caption{Comparison of answer similarity between the static dataset and the AdEval-generated dataset before and after Qwen-7B-Chat contamination, taking BELLE as an example}
  \label{tab:Similarity}
\end{table}

As can be seen from Table~\ref{tab:Similarity}, the answer similarity of the static dataset before and after Qwen-7B-Chat contamination increased by 24.6\%. This indicates that the contamination has a relatively large impact on the static dataset. However, when using the AdEval method, the similarity of answers of Qwen-7B-Chat before and after contamination hardly changed, with only a tiny change of 0.02\%. This result shows that the AdEval method has strong anti-contamination ability in QA tasks.

\subsection{Performance Evaluation}
\label{sec:Performance Evaluation}

\begin{figure}[htbp]
    \centering
  \includegraphics[width=\columnwidth]{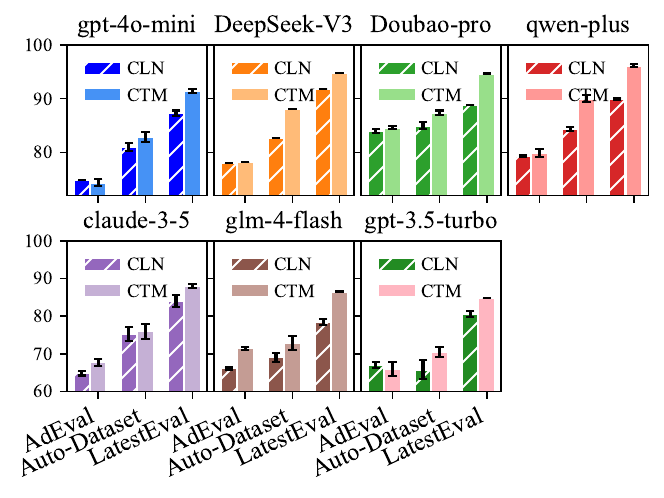}
  \caption {Precision comparison of the datasets generated by AdEaval, Auto-dataset, and Latesteval for MMLU on gpt-3.5-turbo, gpt-4o-mini, DeepSeek-V3, Doubao-pro-32k, qwen-plus, claude-3-5-haiku, glm-4-flash}
  \label{fig:LLMs_test}
\end{figure}


In the previous experiments, open-source LLMs with small parameters were used. In this experiment, closed-source LLMs with large parameters are used to test AdEval, Auto-dataset and Latesteval.
 
Since we can't fine-tune the LLMs in Figure~\ref{fig:LLMs_test}, we add corresponding static-data questions to the prompts to simulate the fine-tuning of data contamination. CLN means testing only with dynamic-data questions, and CTM means adding corresponding static-data questions to the prompts of dynamic-data questions. As can be seen from Figure~\ref{fig:LLMs_test}, the data generated by AdEval has the highest complexity on each model, and it is also hardly affected by static data.

Due to the strict security settings of glm-4-flash, approximately 5\% of the dynamically generated questions were refused to be answered, resulting in certain fluctuations in its test results.

\subsection{Data Reconstruction}

\begin{figure}[htbp]
  \centering
  \includegraphics[width=\columnwidth]{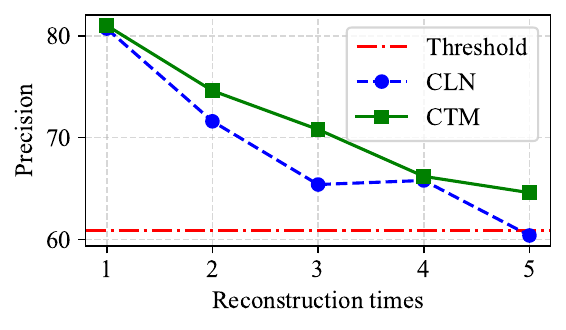}
  \caption{Example of question complexity control in the data reconstruction process of dynamically generated datasets by AdEval, using the MMLU dataset as an example.}
  \label{fig:Reconstruction_Complexity}
\end{figure}



The experiment takes the one based on the MMLU dataset as an example and uses Qwen2-7B-Chat to demonstrate that AdEval controls problem complexity through Data Reconstruction.   In the experiment, AdEval gradually adjusts the complexity of the generated data through multiple rounds of data reconstruction, making it ultimately basically consistent with the complexity of static datasets.   In Figure~\ref{fig:Reconstruction_Complexity}, the dashed line represents the change in the precision of the dynamic dataset in the clean model (CLN);   the solid line represents the change in the precision of the dynamic dataset in the contaminated model (CTM);   the horizontal dashed line represents the benchmark precision of the static data on the clean model;   the abscissa represents the number of reconstructions, and the ordinate represents the precision.   It can be seen from Figure~\ref{fig:Reconstruction_Complexity} that as the number of data reconstruction rounds increases, regardless of whether the model is contaminated or not, the complexity of the data generated by AdEval can gradually increase (the lower the precision, the higher the problem complexity).   Eventually, the precision of the dataset generated by AdEval is basically consistent with the benchmark precision of the static data (with an precision gap within 5\%).


\subsection{Bloom's Cognitive Hierarchy}


\begin{figure}[htbp]
  \centering
  \includegraphics[width=\linewidth]{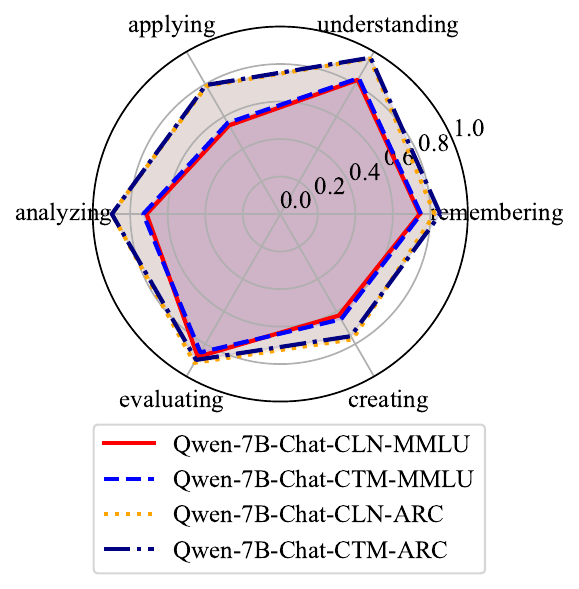}
  \caption{Bloom's Cognitive Hierarchy Radar Chart of Dynamically Generated Datasets by AdEval on MMLU and ARC-Challenge}
  \label{fig:Radar_Bloom}
\end{figure}


This experiment demonstrates that AdEval generates questions of Bloom's cognitive hierarchy, these questions were tested for precision on Qwen-7B-Chat before and after data contamination, and then plotted into a Radar Chart as shown in Figure~\ref{fig:Radar_Bloom}. Questions of Bloom's cognitive hierarchy are generated based on two datasets, MMLU and ARC-Challenge. There are two overlapping radar charts, inner and outer, in the Figure~\ref{fig:Radar_Bloom}. The two overlapping hexagons on the inner side are the precision radar charts of MMLU, and the two overlapping hexagons on the outer side are the precision radar charts of ARC-Challenge. The overlapping radar charts represent the precision of the models before and after contamination, illustrating that the questions of Bloom's cognitive hierarchy generated by AdEval are hardly affected by data contamination. In addition, it can also be seen that the questions of the two datasets are more difficult at the "creating" and "applying" levels, while the questions are less difficult at the "understanding" and "evaluating" levels.


\subsection{Vote Check}



In this experiment, quality checks are conducted on the datasets generated by AdEval, Auto-Dataset, and LatestEval through Vote Check. Three LLMs (Doubao-pro-32k, Qwen-plus, and GPT-4o-mini) are used for voting judgment. The specific rule is that if at least two of the three LLMs judge it as passing, the answer to the question is considered correct; otherwise, the answer is judged to be incorrect. 

Based on MMLU, the error rates of vote checks on the datasets generated by AdEval, Auto-Dataset, and LatestEval are 5.8\%, 9.0\%, and 7.5\% respectively. It can be seen that AdEval has the lowest error rate, indicating that AdEval is superior to the Auto-Dataset and LatestEval methods in terms of correctness in question generation. Meanwhile, through vote checks, this paper identifies incorrect questions, reconstructs them, and further reduces the error rate.


\section{Conclusion and Discussion}


The AdEval method proposed in this paper has achieved remarkable results in addressing the data contamination issue in LLMs evaluation.    By dynamically generating evaluation dataset consistent with static dataset, AdEval not only effectively mitigates the impact of data contamination on LLMs evaluation but also ensures that the generated dataset maintains the same distribution as the static dataset through precise control of question complexity.    Experimental results show that AdEval outperforms the Auto-Dataset and LatestEval methods on the MMLU and ARC-Challenge datasets. Particularly in terms of generated data complexity control and ensuring generation quality, AdEval demonstrates significant advantages.

Future research could further explore the scalability and diversity of the AdEval method, particularly through validation and application on more types of datasets.


\bibliography{aaai2026}


\section{Prompts}
\label{sec:Prompts}

\subsection{Knowledge Point Extraction - Prompt}
\label{sec:Knowledge Point Extraction - Prompt}
You are a professional expert in summarizing knowledge points from questions. Please summarize the relevant knowledge points from the given multiple-choice question.

Requirements:

1. Provide professional knowledge points related to the question.

2. Focus on summarizing broad concepts or categories of knowledge related to the question, rather than specific detailed explanations.

3. Strictly follow the format below and do not output any other content:

[“knowledge1”, “knowledge2” , “knowledge3”, …]

\{\{few-shot\}\}

Please summarize the following multiple-choice question:

\{\{choiceQ\}\}

\hrulefill

The term \{\{few-shot\}\} refers to few-shot learning, with specific details provided in Appendix ~\ref{sec:Knowledge Point Extraction - Few-Shot}. \{\{choiceQ\}\} refers to the questions in the dataset. Both of these are embedded in the prompt and provided as input to the large language model.

\subsection{Main Idea Extraction - Prompt}
\label{sec:Main Idea Extraction - Prompt}

You are a professional expert in summarizing the main idea of questions. Please summarize the main idea based on the input question content.
Requirements:

1. Provide the main idea related to the question, covering the background of the question and the core of the answer.

2. Ensure the main idea is concise and clear, avoiding redundant expressions.

3. Output only the main idea without any additional content.

\{\{few-shot\}\}

Please summarize the main idea of the following question:

\{\{choiceQ\}\}

\subsection{Online Search - Prompt}
\label{sec:Online Search - Prompt}

You are an expert in explaining knowledge points. Based on the provided question, knowledge point, and main idea, use online resources to search for and provide a detailed explanation closely related to the core of the knowledge point. Please strictly follow these requirements for your output:

1. Focus on the Knowledge Point: The explanation must revolve tightly around the input knowledge point. While the question and main idea can provide background or supplementary context, the knowledge point must remain the core focus.  

2. Concise Paragraph: Present the explanation in a single, clear, and coherent paragraph. Avoid excessive length or overly brief statements, and do not include additional structuring or embellishments.  
3. Direct and Comprehensive: Ensure the explanation is accurate, logically sound, and centered on the topic, avoiding irrelevant information or ambiguous statements.

\{\{few-shot\}\}

Please search online and provide a single-paragraph detailed explanation of the knowledge point, referencing the question and main idea where relevant.

Actual Input Question:

\{\{choiceQ\}\}

Actual Input Knowledge Point:

\{\{kn\}\}

Actual Input Main Idea:

\{\{purport\}\}

\hrulefill

\{\{kn\}\} and \{\{purport\}\} represent the knowledge points and main ideas extracted in ~\ref{sec:Knowledge Point Extraction - Prompt} and ~\ref{sec:Main Idea Extraction - Prompt}.

\subsection{Question Design - Prompt}
\label{sec:Question Design - Prompt}

You are a professional question designer specializing in creating challenging multiple-choice questions. Based on the provided reference question, knowledge points, main ideas, and text content, please generate one multiple-choice question and its answer. Ensure the question's difficulty is not lower than the reference question, and preferably slightly higher. The question should test students' in-depth understanding, detailed analysis, and comprehensive judgment skills. The multiple-choice question must meet the following requirements:

1. High Relevance: The question, correct option, and answer must be directly based on the provided text and closely related to the main idea.

2. Consistent or Higher Difficulty: The complexity of the question must match or exceed that of the reference question. Avoid simple factual recitations; instead, require students to achieve a deep understanding and analyze details to deduce the correct answer.

3. Reasonable and Distinguishable Options: The options must be logically clear, avoiding overly obvious correct answers. Include at least one strong distractor to test detailed reading and reasoning skills.

4. Avoid external dependency: The question and its options must be entirely based on the provided text content and main idea. If quoting content is necessary, directly embed the relevant text in the question rather than using references like "according to the passage." Avoid creating questions that depend on context not provided, such as: "Which of the following best describes the purpose of penetration testing as described in the text?"

5. Unique Answer: Ensure only one option is correct, while other options remain plausible but subtly incorrect or mismatched with the main idea and text content.

6. Deep Understanding: The question can involve inference, comparison, method analysis, application of key concepts, etc., to increase the difficulty level.

7. Innovative Question Stem: The question format should differ significantly from the reference question, showcasing creativity while maintaining a rigorous and professional style.

8. Different Focus: The question’s focus should differ from the reference question. For example, if the reference question emphasizes concept definition, the generated question could focus on method application, key detail comparison, pros and cons analysis, or judgment in practical scenarios.

9. Focus on knowledge points: The question must center around the provided knowledge points, with its core content closely aligned with the main idea.

10. Strict Output Format: The output must strictly follow the JSON structure below, without any additional content:

\begin{verbatim}
[
  {
    "Question": "Question content",
    "A": "Option content",
    "B": "Option content",
    "C": "Option content",
    "D": "Option content",
    "Answer": "A/B/C/D"
  }
]
\end{verbatim}

\{\{few-shot\}\}

Below are the provided reference question, knowledge points, main idea, and text content. Please generate a multiple-choice question and answer that meets the above requirements. Remember, do not make the question similar to the reference question, but ensure the difficulty level is consistent with or even higher than the reference question.

Reference question:

\{\{choiceQ\}\}

Knowledge Point:

\{\{kn\}\}

Main Idea:

\{\{purport\}\}

Text Content:

\{\{KNexplain\}\}

\hrulefill

\{\{KNexplain\}\} refers to the detailed explanations of the knowledge points extracted in ~\ref{sec:Online Search - Prompt}.

\subsection{Bloom’s Cognitive Hierarchy Design - Prompt}
\label{sec:Bloom’s Cognitive Hierarchy Design - Prompt}
You are a professional question designer specializing in creating challenging multiple-choice questions. Based on the provided reference question, knowledge points, main ideas, and text content, please generate six multiple-choice questions and answers corresponding to Bloom's six levels of cognitive learning. Ensure the question's difficulty is not lower than the reference question, and preferably slightly higher. The question should test students' in-depth understanding, detailed analysis, and comprehensive judgment skills.

Let me first explain what Bloom's six cognitive levels are:

Introduction to Bloom's Six Cognitive Levels:

Level 1: Remembering

This involves recognizing and recalling concepts and knowledge, storing them in the brain, and retrieving them when needed. Examples include memorizing vocabulary, poetry, or definitions. This level involves identifying concrete or abstract knowledge and can pertain to factual, conceptual, procedural, or metacognitive knowledge. While mechanical, it is a foundational step for learning and solving more complex problems.

Level 2: Understanding

Understanding means grasping the essence of information or knowledge, though not deeply or thoroughly—just a preliminary understanding. When learners connect "new" knowledge to existing knowledge, they achieve understanding. Examples include the classic Feynman technique, where teaching others reinforces personal understanding. Understanding occurs when new information integrates with existing cognitive frameworks.

Understanding includes translating, interpreting, and inferring:

Interpreting involves explaining or summarizing information in one's own words.
Translating involves expressing learned content in different forms, such as using diagrams to explain a concept.

Inferring involves predicting trends or developments based on learned knowledge.

Level 3: Applying

This level focuses on applying learned concepts, principles, or rules in new situations or solving real-world problems. It involves correctly using abstract ideas in appropriate contexts without explicit problem-solving instructions. Applications rely on remembering and understanding as foundational, including applying concepts, principles, methods, and theories, such as using the Pythagorean theorem to solve geometry problems.

Level 4: Analyzing

Analyzing involves breaking down complex knowledge into its components and understanding the relationships between them. It requires deconstructing material to identify essential elements and their relationships, making the material's organization and structure clearer. Tasks include identifying, analyzing, and recognizing the interrelationships and structure among concepts.

For instance, discovering how Factor A changes alongside Factor B by analyzing correlations, causal relationships, mediating variables, and moderators.

Level 5: Evaluating

This level involves making judgments about the value of something based on rational and persuasive reasoning rather than intuition or observation. It requires synthesizing internal and external evidence and making objective, well-supported assessments. For example, legal debates where both sides cite laws to support their arguments exemplify evaluation.

Level 6: Creating

This involves reorganizing learned knowledge or generating new information to form a new whole, such as proposing new ideas, solutions, or designs. It requires analyzing and synthesizing components to form a new structure. Creativity emphasizes innovation and problem-solving, breaking conventional thought patterns and achieving the ultimate learning goal.

Creating is the highest level as it challenges learners to develop new knowledge structures, highlighting innovation and originality.

Multiple-Choice Question Requirements:

1. High Relevance: The question, correct option, and answer must be directly based on the provided text and closely related to the main idea.

2. Consistent or Higher Difficulty: The complexity of the question must match or exceed that of the reference question. Avoid simple factual recitations; instead, require students to achieve a deep understanding and analyze details to deduce the correct answer.

3. Reasonable and Distinguishable Options: The options must be logically clear, avoiding overly obvious correct answers. Include at least one strong distractor to test detailed reading and reasoning skills.

4. Avoid external dependency: The question and its options must be entirely based on the provided text content and main idea. If quoting content is necessary, directly embed the relevant text in the question rather than using references like "according to the passage." Avoid creating questions that depend on context not provided, such as: "Which of the following best describes the purpose of penetration testing as described in the text?"

5. Unique Answer: Ensure only one option is correct, while other options remain plausible but subtly incorrect or mismatched with the main idea and text content.

6. Focus on knowledge points: The question must center around the provided knowledge points, with its core content closely aligned with the main idea.

7. Coverage of Levels: Each question must correspond to a single cognitive level, covering all six levels of Bloom's taxonomy, with one question per level. A total of six questions is required.  

8. Strict Output Format: The output must strictly follow the JSON structure below, without any additional content:
\begin{verbatim}
[
  {
    "Layer": "Cognitive Level",
    "Question": "Question content",
    "A": "Option content",
    "B": "Option content",
    "C": "Option content",
    "D": "Option content",
    "Answer": "A/B/C/D"
  }
  ...
]
\end{verbatim}

\{\{few-shot\}\}

Below are the provided reference question, knowledge points, main idea, and text content. Please generate Bloom's six cognitive-level multiple-choice questions that meets the above requirements. Remember, do not make the question similar to the reference question, but ensure the difficulty level is consistent with or even higher than the reference question.

Reference question:

\{\{choiceQ\}\}

Knowledge Point:

\{\{kn\}\}

Main Idea:

\{\{purport\}\}

Text Content:

\{\{KNexplain\}\}

\subsection{Complexity Control - Prompt}
\label{sec:Complexity Control - Prompt}

You are a professional question reconstruction expert, specializing in enhancing question complexity and designing challenging multiple-choice questions.

Please redesign a multiple-choice question based on the provided original question, current question, knowledge points, main idea, and text content. The goal is to improve students' in-depth understanding, detailed analysis, and comprehensive judgment skills.

Multiple-Choice Question Design Requirements:

1. High Relevance: The stem, correct option, and distractors must be strictly based on the provided text content. Ensure the question is closely related to the main idea, and the answer can be explicitly supported by the text.

2. Increased Difficulty: The question should be more complex than the current one. Avoid simple factual repetition; instead, require students to deeply understand, analyze details, and make connections to arrive at the correct answer.

3. Reasonable and Distinguishable Options: The options must be logically clear, avoiding overly obvious correct answers. Include at least one strong distractor to test detailed reading and reasoning skills.

4. Avoid external dependency: The question and its options must be entirely based on the provided text content and main idea. If quoting content is necessary, directly embed the relevant text in the question rather than using references like "according to the passage." Avoid creating questions that depend on context not provided, such as: "Which of the following best describes the purpose of penetration testing as described in the text?"

5. Unique Answer: Ensure only one option is correct, while other options remain plausible but subtly incorrect or mismatched with the main idea and text content.

6. Deep Understanding: The question can involve inference, comparison, method analysis, application of key concepts, etc., to increase the difficulty level.

7. Innovative Question Stem: The question format should differ significantly from the original question, showcasing creativity while maintaining a rigorous and professional style.

8. Different Focus: The question’s focus should differ from the original question. For example, if the original question emphasizes concept definition, the generated question could focus on method application, key detail comparison, pros and cons analysis, or judgment in practical scenarios.

9. Focus on knowledge points: The question must center around the provided knowledge points, with its core content closely aligned with the main idea.
10. Strict Output Format: The output must strictly follow the JSON structure below, without any additional content:
\begin{verbatim}
[
  {
    "Question": "Question content",
    "A": "Option content",
    "B": "Option content",
    "C": "Option content",
    "D": "Option content",
    "Answer": "A/B/C/D"
  }
]
\end{verbatim}

\{\{few-shot\}\}

The following includes the original question, current question, knowledge points, main idea, and text content. Please generate a multiple-choice question that meets the above requirements and provide an answer. Be sure not to make it similar to the original question, and ensure that the difficulty of the newly generated question is higher than that of the current question.

Original question:

\{\{choiceQ\}\}

current question:

\{\{choiceQCurrent\}\}

Knowledge Point:

\{\{kn\}\}

Main Idea:

\{\{purport\}\}

Text Content:

\{\{KNexplain\}\}

\hrulefill

\{\{choiceQCurrent\}\} refers to the questions dynamically generated in ~\ref{sec:Question Design - Prompt} and ~\ref{sec:Bloom’s Cognitive Hierarchy Design - Prompt}.

\subsection{Quality Control - Prompt}
\label{sec:Quality Control - Prompt}
\subsubsection{Question Answer Checking Prompt}
\label{Question Answer Checking Prompt}
You are a professional multiple-choice question proofreading expert, specializing in verifying the accuracy of questions and their answers.

I will provide a multiple-choice question and its answer.
Please strictly follow the rules below to make your judgment and respond only in numeric form:

1. If the answer is incorrect, respond with "0".

2. If the answer is correct, respond with "1".

3. Your response must be either 0/1, with no additional content.

Example Input:

Which of the following best describes the balance the Supreme Court has struck between the establishment clause and the free-exercise clause? A Freedom of speech is protected except in certain situations, such as yelling "fire" in a crowded theater. B Once a church has been recognized by the federal government, its tax-exempt status can never be revoked. C Once Congress has created an administrative agency, that agency can be dissolved only by a constitutional amendment. D State-sponsored prayer during school hours is prohibited, but voluntary prayer by student groups before school is allowed. 

Answer:D

Example Output:

0

Below is a multiple-choice question and its answer. Please judge according to the rules and respond only with 0/1:

\{\{choiceQ\}\}

\section{Few-Shot}
\label{sec:Few-Shot}

\subsection{Knowledge Point Extraction - Few-Shot}
\label{sec:Knowledge Point Extraction - Few-Shot}

Example Input1:

"Question: Which of the following regular expressions is equivalent to (describes the same set of strings as) (a* + b)*(c + d)? 

A a*(c + d)+ b(c + d). 

B a*(c + d)* + b(c + d)*. 

C a*(c + d)+ b*(c + d). 

D (a + b)*c +(a + b)*d

Answer:D"

Example Output1:

["Regular expressions and their operators", "Concatenation in regular expressions", "Union operator (+) in regular expressions", "Kleene star (*) in regular expressions", "Equivalence of regular expressions", "Pattern matching with regular expressions"]

Example Input2:

"Question:  A certain pipelined RISC machine has 8 general-purpose registers R0, R1, . . . , R7 and supports the following operations.
ADD Rs1, Rs2, Rd Add Rs1 to Rs2 and put the sum in Rd

MUL Rs1, Rs2, Rd Multiply Rs1 by Rs2 and put the product in Rd

An operation normally takes one cycle; however, an operation takes two cycles if it produces a result required by the immediately following operation in an operation sequence. Consider the expression AB + ABC + BC, where variables A, B, C are located in registers R0, R1, R2. If the contents of these three registers must not be modified, what is the minimum number of clock cycles required for an operation sequence that computes the value of AB + ABC + BC?"

A 5  B 6  C 7  D 8

Answer:B"

Example Output2:

["RISC architecture and instruction pipelining", "Data hazards and their impact on instruction execution", "Register allocation and usage in assembly language", "Instruction scheduling to minimize clock cycles", "Arithmetic operations in assembly language"]

\subsection{Main Idea Extraction - Few-Shot}
\label{sec:Main Idea Extraction - Few-Shot}

Example Input1:

"Question: Which of the following regular expressions is equivalent to (describes the same set of strings as) (a* + b)*(c + d)? 

A a*(c + d)+ b(c + d). 

B a*(c + d)* + b(c + d)*. 

C a*(c + d)+ b*(c + d). 

D (a + b)*c +(a + b)*d

Answer:D"

Example Output1:

Finding the regular expression equivalent to (a* + b)*(c + d).

Example Input2:

"Question:  A certain pipelined RISC machine has 8 general-purpose registers R0, R1, . . . , R7 and supports the following operations.
ADD Rs1, Rs2, Rd Add Rs1 to Rs2 and put the sum in Rd
MUL Rs1, Rs2, Rd Multiply Rs1 by Rs2 and put the product in Rd
An operation normally takes one cycle; however, an operation takes two cycles if it produces a result required by the immediately following operation in an operation sequence. Consider the expression AB + ABC + BC, where variables A, B, C are located in registers R0, R1, R2. If the contents of these three registers must not be modified, what is the minimum number of clock cycles required for an operation sequence that computes the value of AB + ABC + BC?" 

A 5 B 6  C 7  D 8

Answer:B"

Example Output2:

Determining the minimum number of clock cycles required to compute the expression \( AB + ABC + BC \) on a pipelined RISC machine, considering operation latencies, data dependencies, and constraints on register modification.

\subsection{Online Search - Few-Shot}
\label{sec:Online Search - Few-Shot}
Example Input1:

Question: 

Which of the following regular expressions is equivalent to (describes the same set of strings as) (a* + b)*(c + d)? 

A a*(c + d)+ b(c + d). 

B a*(c + d)* + b(c + d)*. 

C a*(c + d)+ b*(c + d). 

D (a + b)*c +(a + b)*d

Answer:D

Knowledge Point: 

Regular expressions and their operators

Main Idea: 
Finding the regular expression equivalent to (a* + b)*(c + d).

Example Output1:

Regular expressions are sequences of characters that define search patterns, primarily used for pattern matching within strings. The fundamental operators in regular expressions include concatenation (placing expressions sequentially), alternation (denoted by `|` or `+`, representing a choice between expressions), and the Kleene star (denoted by `*`, indicating zero or more repetitions of the preceding element). Understanding these operators is crucial for constructing expressions that accurately represent specific string patterns. For instance, the expression `(a|b)*` denotes any combination of the characters 'a' and 'b', including the empty string, while `a*` matches any number of consecutive 'a' characters, including none. Mastery of these operators enables the creation of complex patterns for efficient text processing and matching tasks.  

Example Input2:

Question:  A certain pipelined RISC machine has 8 general-purpose registers R0, R1, . . . , R7 and supports the following operations.
ADD Rs1, Rs2, Rd Add Rs1 to Rs2 and put the sum in Rd
MUL Rs1, Rs2, Rd Multiply Rs1 by Rs2 and put the product in Rd
An operation normally takes one cycle; however, an operation takes two cycles if it produces a result required by the immediately following operation in an operation sequence. Consider the expression AB + ABC + BC, where variables A, B, C are located in registers R0, R1, R2. If the contents of these three registers must not be modified, what is the minimum number of clock cycles required for an operation sequence that computes the value of AB + ABC + BC?" 

A 5  B 6  C 7  D 8

Answer:B

Knowledge Point: 

Register allocation and usage in assembly language

Main Idea: 

Determining the minimum number of clock cycles required to compute the expression \( AB + ABC + BC \) on a pipelined RISC machine, considering operation latencies, data dependencies, and constraints on register modification.

Example Output2:

In assembly language programming, efficient register allocation and usage are crucial for optimizing performance, especially in pipelined RISC architectures. Registers are limited, high-speed storage locations within the CPU that hold data for immediate processing, reducing the need to access slower main memory. Effective register allocation involves assigning these registers to frequently used variables and intermediate results, minimizing memory access and enhancing execution speed. This process can be complex due to the limited number of registers and the need to manage data dependencies and instruction latencies. Techniques such as graph coloring are often employed to allocate registers efficiently by modeling interference between variables and ensuring that no two simultaneously live variables share the same register. Additionally, understanding the specific roles and constraints of registers in a given architecture is essential, as certain instructions may require operands to reside in specific registers. By carefully managing register allocation and usage, assembly language programmers can significantly improve the efficiency and performance of their code.

\subsection{Question Design - Few-Shot}
\label{sec:Question Design - Few-Shot}
Example Input1:

Reference question:

Which of the following regular expressions is equivalent to (describes the same set of strings as) (a* + b)*(c + d)? 

A a*(c + d)+ b(c + d). 

B a*(c + d)* + b(c + d)*. 

C a*(c + d)+ b*(c + d). 

D (a + b)*c +(a + b)*d

Answer:D

Knowledge Point:

Regular expressions and their operators

Main Idea:

Finding the regular expression equivalent to (a* + b)*(c + d).

Text Content:

Regular expressions are sequences of characters that define search patterns, primarily used for pattern matching within strings. The fundamental operators in regular expressions include concatenation (placing expressions sequentially), alternation (denoted by `|` or `+`, representing a choice between expressions), and the Kleene star (denoted by `*`, indicating zero or more repetitions of the preceding element). Understanding these operators is crucial for constructing expressions that accurately represent specific string patterns. For instance, the expression `(a|b)*` denotes any combination of the characters 'a' and 'b', including the empty string, while `a*` matches any number of consecutive 'a' characters, including none. Mastery of these operators enables the creation of complex patterns for efficient text processing and matching tasks.

Example Output1:

[
  \{
  
    "Question": "Which of the following regular expressions correctly represents all strings that start with zero or more repetitions of either 'x' or 'y', followed by exactly two repetitions of either 'm' or 'n', and end with a single occurrence of either 'p' or 'q'?",
    
    "A": "$(x|y)*((m|n)(m|n))(p|q)$",
    
    "B": "$((x|y)*(m|n))^2(p|q)$",
    
    "C": "$(x* + y*)((m + n)(m + n))(p + q)$",
    
    "D": "$(x|y)^2((m|n)(m|n))(p|q)$",
    
    "Answer": "A"
    
  \}
]

Example Input2:

Reference question:

A certain pipelined RISC machine has 8 general-purpose registers R0, R1, . . . , R7 and supports the following operations.
ADD Rs1, Rs2, Rd Add Rs1 to Rs2 and put the sum in Rd
MUL Rs1, Rs2, Rd Multiply Rs1 by Rs2 and put the product in Rd
An operation normally takes one cycle; however, an operation takes two cycles if it produces a result required by the immediately following operation in an operation sequence. Consider the expression AB + ABC + BC, where variables A, B, C are located in registers R0, R1, R2. If the contents of these three registers must not be modified, what is the minimum number of clock cycles required for an operation sequence that computes the value of AB + ABC + BC?" 

A 5  B 6  C 7  D 8

Answer:B

Knowledge Point:

Register allocation and usage in assembly language

Main Idea:

Determining the minimum number of clock cycles required to compute the expression \( AB + ABC + BC \) on a pipelined RISC machine, considering operation latencies, data dependencies, and constraints on register modification.

Text Content:

In assembly language programming, efficient register allocation and usage are crucial for optimizing performance, especially in pipelined RISC architectures. Registers are limited, high-speed storage locations within the CPU that hold data for immediate processing, reducing the need to access slower main memory. Effective register allocation involves assigning these registers to frequently used variables and intermediate results, minimizing memory access and enhancing execution speed. This process can be complex due to the limited number of registers and the need to manage data dependencies and instruction latencies. Techniques such as graph coloring are often employed to allocate registers efficiently by modeling interference between variables and ensuring that no two simultaneously live variables share the same register. Additionally, understanding the specific roles and constraints of registers in a given architecture is essential, as certain instructions may require operands to reside in specific registers. By carefully managing register allocation and usage, assembly language programmers can significantly improve the efficiency and performance of their code. 

Example Output2:

[
  \{
  
    "Question": "Given the same pipelined RISC machine and the expression \( AB + ABC + BC \), where A, B, and C are in registers R0, R1, and R2 respectively, and assuming no register modification is allowed, which sequence of operations would result in the minimum number of clock cycles, and what is that minimum number?",
    
    "A": "Perform MUL R0, R1, R3; MUL R1, R2, R4; ADD R3, R4, R5; ADD R5, R2, R6; 7 cycles",
    
    "B": "Perform MUL R0, R1, R3; ADD R3, R1, R4; MUL R1, R2, R5; ADD R4, R5, R6; 7 cycles",
    
    "C": "Perform MUL R0, R1, R3; MUL R3, R2, R4; ADD R3, R2, R5; ADD R4, R5, R6; 6 cycles",
    
    "D": "Perform MUL R0, R1, R3; MUL R1, R2, R4; ADD R3, R4, R5; ADD R5, R1, R6; 6 cycles",
    
    "Answer": "C"
    
  \}
]

Finally, remember not to output in the following format (do not begin with \begin{verbatim}```\end{verbatim}json and end with \begin{verbatim}```\end{verbatim}).

\begin{verbatim}```\end{verbatim}json
[
  \{
    "Question": "Which of the following regular expressions correctly represents all strings that can be formed by zero or more repetitions of the strings made up of 'a' or 'b', followed by exactly one occurrence of either 'c' or 'd'?",
    
    "A": "$(a|b)*(c|d)$",
    
    "B": "$(a*b*+b*a*)(c|d)$",
    
    "C": "$(a+b)*(c|d)+$",
    
    "D": "$((a+b)*c) + ((a+b)*d)$",
    
    "Answer": "A"
    
  \}
]
\begin{verbatim}```\end{verbatim}

\subsection{Bloom's Cognitive Hierarchy Question Design - Few-Shot}

\label{Bloom's Cognitive Hierarchy Question Design - Few-Shot}

Example Input1:

Reference question:

When trying to construct a regular expression to match strings that start with an optional single occurrence of either 'e' or 'f', followed by zero or more repetitions of a group containing either 'g' or 'h' and then either 'i' or 'j', and end with exactly three repetitions of either 'k' or 'l', which of the following options would be the correct regular expression?  

A $(e|f)?((g|h)*(i|j))(k|l)^3$

B $((e|f)+(g|h))*(i|j)(k|l)^3$ 

C $(e|f)(g|h)*(i|j)^3(k|l)^3$ 

D $(e|f)*((g|h)+(i|j))(k|l)^3$ 

Answer: A

Knowledge Point:

Regular expressions and their operators

Main Idea:

Finding the regular expression equivalent to (a* + b)*(c + d).

Text Content:

Regular expressions are sequences of characters that define search patterns, primarily used for pattern matching within strings. The fundamental operators in regular expressions include concatenation (placing expressions sequentially), alternation (denoted by `|` or `+`, representing a choice between expressions), and the Kleene star (denoted by `*`, indicating zero or more repetitions of the preceding element). Understanding these operators is crucial for constructing expressions that accurately represent specific string patterns. For instance, the expression `(a|b)*` denotes any combination of the characters 'a' and 'b', including the empty string, while `a*` matches any number of consecutive 'a' characters, including none. Mastery of these operators enables the creation of complex patterns for efficient text processing and matching tasks.

Example Output1:

[
  \{
  
    "Layer": "Remembering",
    "Question": "Which symbol in regular expressions denotes zero or more repetitions of the preceding element?",
    
    "A": "?", "B": "*", "C": "+", "D": ".",
    
    "Answer": "B"
    
  \},
  \{
    "Layer": "Understanding",
    
    "Question": "In the context of regular expressions, what does the expression `(a|b)*` represent?",
    
    "A": "Any string that contains only 'a' repeated any number of times",
    
    "B": "Any string that contains only 'b' repeated any number of times",
    
    "C": "Any combination of 'a' and 'b', including the empty string",
    
    "D": "Any string that starts with 'a' or 'b' and ends with 'a' or 'b'",
    
    "Answer": "C"
    
  \},
  
  \{
    "Layer": "Applying",
    
    "Question": "Given the regular expression `(a* + b)*(c + d)`, which of the following regular expressions correctly applies the same pattern?",
    
    "A": "$(a|b)*(c|d)$",
    
    "B": "$(a+b)*(c+d)$",
    
    "C": "$(a*b)*(c+d)$",
    
    "D": "$(a+b)*(c|d)$",
    
    "Answer": "D"
    
  \},
  
  \{
  
    "Layer": "Analyzing",
    
    "Question": "How does the regular expression `(a|b)*` differ from `a*` in terms of the possible strings it can match?",
    
    "A": "They match exactly the same set of strings",
    
    "B": "`(a|b)*` can match strings with 'b', while `a*` cannot",
    
    "C": "`a*` can match strings with 'b', while `(a|b)*` cannot",
    
    "D": "They both match strings with any number of 'a's, including 'b's",
    
    "Answer": "B"
    
  \},
  
  \{
  
    "Layer": "Evaluating",
    
    "Question": "Considering the regular expression `(a* + b)*(c + d)`, evaluate which of the following statements best justifies the use of the alternation operator `|` over the concatenation operator `+` in this context?",
    
    "A": "The `|` operator allows for a choice between 'a' and 'b', which is necessary for matching a pattern that could start with either 'a' or 'b'",
    
    "B": "The `+` operator is used for one or more repetitions, which is not suitable for optional elements",
    
    "C": "The `|` operator is more efficient for matching a single character",
    
    "D": "The `|` operator is used to match either 'a' or 'b', but it is not necessary for the pattern to be valid",
    
    "Answer": "A"
    
  \},
  
  \{
  
    "Layer": "Creating",
    
    "Question": "If you were to create a new regular expression based on the pattern `(a* + b)*(c + d)` that matches a string starting with zero or more 'a's, followed by one or more 'b's, and ending with either 'c' or 'd', what would that regular expression be?",
    
    "A": "(a*b+)(c|d)",
    
    "B": "(a+b*)(c|d)",
    
    "C": "(a* + b+)(c|d)",
    
    "D": "(a+b*)(c+d)",
    
    "Answer": "C"
    
  \}
]

Finally, remember not to output in the following format (do not begin with \begin{verbatim}```\end{verbatim}json and end with \begin{verbatim}```\end{verbatim}).

```json

[
  \{
  
    "Layer": "Remembering",
    
    "Question": "Which symbol in regular expressions denotes zero or more repetitions of the preceding element?",
    
    "A": "?", "B": "*", "C": "+", "D": ".",
    
    "Answer": "B"
  
  \},
] 

\begin{verbatim}```\end{verbatim}

\subsection{Complexity Control - Few-Shot}
\label{sec:Complexity Control - Few-Shot}

Example Input1:

Original question:

Question: Which of the following regular expressions is equivalent to (describes the same set of strings as) (a* + b)*(c + d)? 

A a*(c + d)+ b(c + d). 

B a*(c + d)* + b(c + d)*. 

C a*(c + d)+ b*(c + d). 

D (a + b)*c +(a + b)*d

Answer:D

current question:

Which of the following regular expressions correctly represents all strings that start with zero or more repetitions of either 'x' or 'y', followed by exactly two repetitions of either 'm' or 'n', and end with a single occurrence of either 'p' or 'q'?

A $(x|y)*((m|n)(m|n))(p|q)$

B $((x|y)*(m|n))^2(p|q)$

C $(x* + y*)((m + n)(m + n))(p + q)$ 

D $(x|y)^2((m|n)(m|n))(p|q)$  

Answer:A

Knowledge Point:

Regular expressions and their operators

Main Idea:

Finding the regular expression equivalent to (a* + b)*(c + d).

Text Content:

Regular expressions are sequences of characters that define search patterns, primarily used for pattern matching within strings. The fundamental operators in regular expressions include concatenation (placing expressions sequentially), alternation (denoted by `|` or `+`, representing a choice between expressions), and the Kleene star (denoted by `*`, indicating zero or more repetitions of the preceding element). Understanding these operators is crucial for constructing expressions that accurately represent specific string patterns. For instance, the expression `(a|b)*` denotes any combination of the characters 'a' and 'b', including the empty string, while `a*` matches any number of consecutive 'a' characters, including none. Mastery of these operators enables the creation of complex patterns for efficient text processing and matching tasks.

Example Output1:

[
  \{
    "Question": "When trying to construct a regular expression to match strings that start with an optional single occurrence of either 'e' or 'f', followed by zero or more repetitions of a group containing either 'g' or 'h' and then either 'i' or 'j', and end with exactly three repetitions of either 'k' or 'l', which of the following options would be the correct regular expression? ",
    
    "A": "$(e|f)?((g|h)*(i|j))(k|l)^3$",
    
    "B": "$((e|f)+(g|h))*(i|j)(k|l)^3$",
    
    "C": "$(e|f)(g|h)*(i|j)^3(k|l)^3$",
    
    "D": "$(e|f)*((g|h)+(i|j))(k|l)^3$",
    
    "Answer": "A"
  \}
]

Example Input2:

Original question:

Question:  A certain pipelined RISC machine has 8 general-purpose registers R0, R1, . . . , R7 and supports the following operations.
ADD Rs1, Rs2, Rd Add Rs1 to Rs2 and put the sum in Rd
MUL Rs1, Rs2, Rd Multiply Rs1 by Rs2 and put the product in Rd
An operation normally takes one cycle; however, an operation takes two cycles if it produces a result required by the immediately following operation in an operation sequence. Consider the expression AB + ABC + BC, where variables A, B, C are located in registers R0, R1, R2. If the contents of these three registers must not be modified, what is the minimum number of clock cycles required for an operation sequence that computes the value of AB + ABC + BC?" 

A 5  B 6  C 7  D 8

Answer:B

current question:

Given the same pipelined RISC machine and the expression \( AB + ABC + BC \), where A, B, and C are in registers R0, R1, and R2 respectively, and assuming no register modification is allowed, which sequence of operations would result in the minimum number of clock cycles, and what is that minimum number? 

A Perform MUL R0, R1, R3; MUL R1, R2, R4; ADD R3, R4, R5; ADD R5, R2, R6; 7 cycles 

B Perform MUL R0, R1, R3; ADD R3, R1, R4; MUL R1, R2, R5; ADD R4, R5, R6; 7 cycles 

C Perform MUL R0, R1, R3; MUL R3, R2, R4; ADD R3, R2, R5; ADD R4, R5, R6; 6 cycles 

D Perform MUL R0, R1, R3; MUL R1, R2, R4; ADD R3, R4, R5; ADD R5, R1, R6; 6 cycles   

Answer: C

Knowledge Point:

Register allocation and usage in assembly language

Main Idea:

Determining the minimum number of clock cycles required to compute the expression \( AB + ABC + BC \) on a pipelined RISC machine, considering operation latencies, data dependencies, and constraints on register modification.

Text Content:

In assembly language programming, efficient register allocation and usage are crucial for optimizing performance, especially in pipelined RISC architectures. Registers are limited, high-speed storage locations within the CPU that hold data for immediate processing, reducing the need to access slower main memory. Effective register allocation involves assigning these registers to frequently used variables and intermediate results, minimizing memory access and enhancing execution speed. This process can be complex due to the limited number of registers and the need to manage data dependencies and instruction latencies. Techniques such as graph coloring are often employed to allocate registers efficiently by modeling interference between variables and ensuring that no two simultaneously live variables share the same register. Additionally, understanding the specific roles and constraints of registers in a given architecture is essential, as certain instructions may require operands to reside in specific registers. By carefully managing register allocation and usage, assembly language programmers can significantly improve the efficiency and performance of their code. 

Example Output2:

[
  \{
  
    "Question": "When computing the expression ( AB + ABC + BC ) on the described pipelined RISC machine, if we consider not only the operation latencies and register modification constraints but also the impact of register allocation strategies like graph coloring on reducing data dependencies, which of the following operation sequences would be the most efficient in terms of minimizing the number of clock cycles while also ensuring the best utilization of registers considering potential interference between variables according to the principles of register allocation in assembly language, and what is that minimum number of cycles?",
    
    "A": "Perform MUL R0, R1, R3; then use graph coloring to reallocate R1 for another operation to avoid interference; MUL R3, R2, R4; ADD R3, R2, R5; ADD R4, R5, R6; 5 cycles",
    
    "B": "First apply graph coloring to allocate registers optimally for the initial MUL operations; perform MUL R0, R1, R3; MUL R1, R2, R4; then handle ADD operations considering data dependencies based on the coloring result; ADD R3, R4, R5; ADD R5, R2, R6; 6 cycles",
    
    "C": "Without using graph coloring, just follow a traditional register allocation based on operand order; perform MUL R0, R1, R3; MUL R3, R2, R4; ADD R3, R2, R5; ADD R4, R5, R6; 7 cycles",
    
    "D": "Use graph coloring but in a suboptimal way that leads to some unnecessary register swaps; perform MUL R0, R1, R3; then swap R1 with another register due to the coloring plan; MUL R1, R2, R4; ADD R3, R4, R5; ADD R5, R2, R6; 8 cycles",
    
    "Answer": "B"
    
  \}
]

Finally, remember not to output in the following format (do not begin with \begin{verbatim}```\end{verbatim}json and end with \begin{verbatim}```\end{verbatim}).

\begin{verbatim}```\end{verbatim}json

[
  \{
    "Question": "When trying to construct a regular expression to match strings that start with an optional single occurrence of either 'e' or 'f', followed by zero or more repetitions of a group containing either 'g' or 'h' and then either 'i' or 'j', and end with exactly three repetitions of either 'k' or 'l', which of the following options would be the correct regular expression? ",
    
    "A": "$(e|f)?((g|h)*(i|j))(k|l)^3$",
    
    "B": "$((e|f)+(g|h))*(i|j)(k|l)^3$",
    
    "C": "$(e|f)(g|h)*(i|j)^3(k|l)^3$",
    
    "D": "$(e|f)*((g|h)+(i|j))(k|l)^3$",
    
    "Answer": "A"
  \}
] 

\begin{verbatim}```\end{verbatim}

\section{Question Design Example Based on Bloom's cognitive hierarchy}
\label{Question Design Example Based on Bloom's cognitive hierarchy}

\textbf{Layer: Remembering}

Question: Which symbol in regular expressions denotes zero or more repetitions of the preceding element?

A: ?     B: *     C: +     D: .

Answer: B

\textbf{Layer: Understanding}

Question: In the context of regular expressions, what does the expression `(a|b)*` represent?

A: Any string that contains only 'a' repeated any number of times 

B: Any string that contains only 'b' repeated any number of times 

C: Any combination of 'a' and 'b', including the empty string

D: Any string that starts with 'a' or 'b' and ends with 'a' or 'b'

Answer: C

\textbf{Layer: Applying}

Question: Given the regular expression `(a* + b)*(c + d)`, which of the following regular expressions correctly applies the same pattern?

A: (a|b)*(c|d)         B: (a+b)*(c+d) 

C: (a*b)*(c+d)       D: (a+b)*(c|d)

Answer: D

\textbf{Layer: Analyzing}

Question: How does the regular expression `(a|b)*` differ from `a*` in terms of the possible strings it can match?

A: They match exactly the same set of strings

B: `(a|b)*` can match strings with 'b', while `a*` cannot

C: `a*` can match strings with 'b', while `(a|b)*` cannot

D: They both match strings with any number of 'a's including 'b's

Answer: B

\textbf{Layer: Evaluating}

Question: Considering the regular expression `(a* + b)*(c + d)`, evaluate which of the following statements best justifies the use of the alternation operator `|` over the concatenation operator `+` in this context?

A: The `|` operator allows for a choice between 'a' and 'b', which is necessary for matching a pattern that could start with either 'a' or 'b'

B: The `+` operator is used for one or more repetitions, which is not suitable for optional elements

C: The `|` operator is more efficient for matching a single character

D: The `|` operator is used to match either 'a' or 'b', but it is not necessary for the pattern to be valid

Answer: A

\textbf{Layer: Creating}

Question: If you were to create a new regular expression based on the pattern `(a* + b)*(c + d)` that matches a string starting with zero or more 'a's, followed by one or more 'b's, and ending with either 'c' or 'd', what would that regular expression be?,

A: (a*b+)(c|d)       B: (a+b*)(c|d)

C: (a* + b+)(c|d)   D: (a+b*)(c+d)

Answer: C

\section{Comparison of examples of dynamic dataset generation methods}

\textbf{Static}

An astronomer observes that a planet rotates faster after a meteorite impact. Which is the most likely effect of this increase in rotation?

A Planetary density will decrease. 

B Planetary years will become longer. 

C Planetary days will become shorter. 

D Planetary gravity will become stronger.

Answer: C

\textbf{AdEval}

A planet experiences a significant increase in its rotation speed due to a series of meteorite impacts. If the planet's initial angular momentum was \(L_{1}\) and its final angular momentum after the impacts is \(L_{2}\), and the initial length of a planetary day was \(T_{1}\) and the final length is \(T_{2}\), which of the following relationships correctly describes the change in angular momentum and the change in the length of the planetary day, assuming no other significant external torques act on the planet? 

A \(L_{2}>L_{1}\) and \(T_{2}<T_{1}\) 

B \(L_{2}<L_{1}\) and \(T_{2}>T_{1}\) 

C \(L_{2}>L_{1}\) and \(T_{2}>T_{1}\) 

D \(L_{2}<L_{1}\) and \(T_{2}<T_{1}\)

Answer: A

\textbf{Auto-Dataset}

An astronaut notices that a celestial body spins more rapidly after a large asteroid collision. What is the most probable consequence of this enhanced rotation? 

A The mass of the celestial body will increase. 

B The seasons on the celestial body will last longer. 

C The length of a day on the celestial body will be reduced. 

D The magnetic field of the celestial body will weaken.

Answer: C

\textbf{LatestEaval}
According to the principle of conservation of angular momentum, when a meteorite impacts a planet and imparts additional angular momentum, what is the most likely immediate effect? 

A The density of the planet increases. 

B The length of the planetary year decreases. 

C The length of the planetary day becomes shorter. 

D The gravity of the planet increases. 

Answer: C

\section{LLaMA-Factory Fine-tunes LLMs}
\begin{figure}[htbp]
  \centering
  \includegraphics[width=\columnwidth]{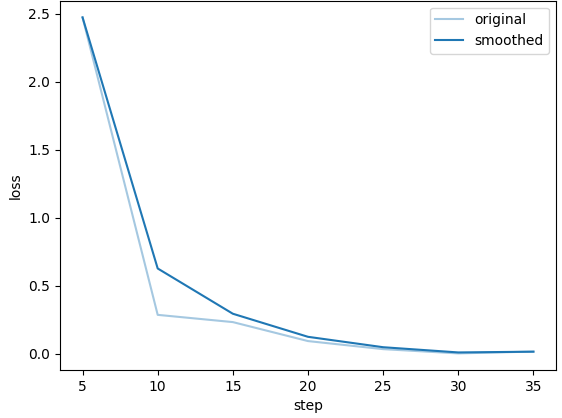}
  \caption {The loss of Qwen2-7B-Chat trained in LLaMA-Factory}
  \label{Qwen2-7B-Chat_training_loss}
\end{figure}

\begin{figure}[htbp]
  \centering
  \includegraphics[width=\columnwidth]{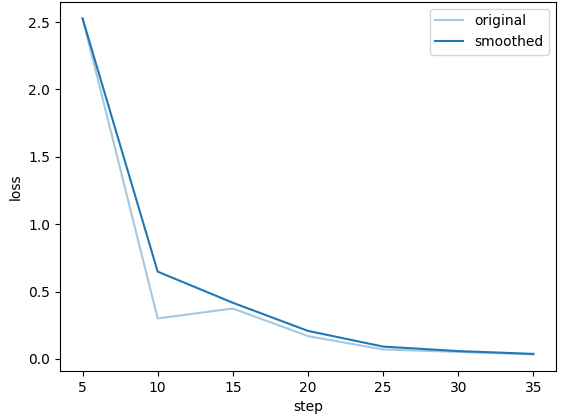}
  \caption {The loss of Llama3-8B-Instruct trained in LLaMA-Factory}
  \label{Llama3-8B-Instruct_training_loss}
\end{figure}

The loss values during the fine-tuning of Qwen2-7B-Chat and Llama3-8B-Instruct using LLaMA-Factory are shown in Figure \ref{Qwen2-7B-Chat_training_loss} and Figure \ref{Llama3-8B-Instruct_training_loss}. It can be seen that due to the small amount of experimental data, the loss values approach 0 after 35 iterations, and at the same time, the loss values decrease at the fastest rate during the first 10 iterations.

\end{document}